\documentclass{article}%
\usepackage{amsmath}
\usepackage{amsfonts}
\usepackage{amssymb}
\usepackage{graphicx}
\usepackage{algorithmic}
\usepackage{algorithm}%
\providecommand{\U}[1]{\protect\rule{.1in}{.1in}}
\newtheorem{theorem}{Theorem}
\newtheorem{definition}[theorem]{Definition}
\begin{document}

\title{Counterfactual explanation of machine learning survival models}
\author{Maxim S. Kovalev and Lev V. Utkin\\Peter the Great St.Petersburg Polytechnic University (SPbPU)\\St.Petersburg, Russia\\e-mail:maxkovalev03@gmail.com, lev.utkin@gmail.com}
\date{}
\maketitle

\begin{abstract}
A method for counterfactual explanation of machine learning survival models is
proposed. One of the difficulties of solving the counterfactual explanation
problem is that the classes of examples are implicitly defined through
outcomes of a machine learning survival model in the form of survival
functions. A condition that establishes the difference between survival
functions of the original example and the counterfactual is introduced. This
condition is based on using a distance between mean times to event. It is
shown that the counterfactual explanation problem can be reduced to a standard
convex optimization problem with linear constraints when the explained
black-box model is the Cox model. For other black-box models, it is proposed
to apply the well-known Particle Swarm Optimization algorithm. A lot of
numerical experiments with real and synthetic data demonstrate the proposed method.


\end{abstract}

\section{Introduction}

Explanation of machine learning models has become an important problem in many
applications. For instance, a lot of medicine machine learning applications
meet a requirement of understanding results provided by the models. A typical
example is when a doctor has to have an explanation of a stated diagnosis in
order to have an opportunity to choose a preferable treatment
\cite{Holzinger-etal-2019}. This implies that the machine learning model
decisions can be trusted and explainable to help machine learning users or
experts to understand the obtained results. One of the obstacles to obtain
explainable decisions is the black-box nature of many models, especially, of
deep learning models, i.e., inputs and outcomes of these models may be known,
but there is no information what features impact on corresponding decisions
provided by the models. A lot of explanation methods have been developed in
order to overcome this obstacle
\cite{Arya-etal-2019,Guidotti-2019,Molnar-2019,Murdoch-etal-2019} and to
explain the model outcomes. The explanation methods can be divided into two
groups: local and global methods. Methods from the first group derive
explanation locally around a test example whereas methods from the second
group try to explain the black-box model on the whole dataset or its part. The
global explanation methods are of a high interest, but many application areas,
especially, medicine, require to understand decisions concerning with a
patient, i.e., it is important to know what features of an example are
responsible for a black-box model prediction. Therefore, this paper focuses on
local explanations.

It is important to note that users of a black-box model are often not
interested why a certain prediction was obtained and what features of an
example led to a decision. They usually aim to understand what could be
changed to get a preferable result by using the black-box model. Such
explanations can be referred to the so-called counterfactual explanations or
counterfactuals \cite{Wachter-etal-2017}, which may be more desirable,
intuitive and useful than \textquotedblleft direct\textquotedblright%
\ explanations (methods based on attributing a prediction to input features).
According to \cite{Molnar-2019}, a counterfactual explanation of a prediction
can be defined as the smallest change to the feature values of an input
original example that changes the prediction to a predefined outcome. There is
a classic example of the loan application rejection
\cite{Molnar-2019,Wachter-etal-2017}, which explicitly explains
counterfactuals. According to this example, a bank rejects an application of a
user for a credit. A counterfactual explanation could be that
\textquotedblleft the credit would have been approved if the user would earn
\$500 more per month and have the credit score 30 points
higher\textquotedblright\ \cite{Molnar-2019,Wachter-etal-2017}.

So far, methods of counterfactual explanations have been applied to
classification and regression problems where a black-box model produces a
point-value outcome for every input example. However, there are a lot of
models, where the outcome is a function. A part of these models solves
problems of survival analysis \cite{Hosmer-Lemeshow-May-2008}, where the
outcome is a survival function (SF) or a cumulative hazard function (CHF). In
contrast to usual classification and regression machine learning models, the
survival models deal with datasets containing a lot of censored data. This
fact complicates the models. A lot of machine learning survival models have
been developed \cite{Lee-Zame-etal-2018,Wang-Li-Reddy-2017,Zhao-Feng-2019} due
to importance the survival models in many applications, including reliability
of complex systems, medicine, risk analysis, etc. In spite of some strong
assumptions, the semi-parametric Cox proportional hazards model
\cite{Cox-1972} remains one of the most popular one. It is one of the models
establishing an explicit relationship between the covariates and the
distribution of survival times. The Cox model assumes a linear combination of
the example covariates. On the one hand, this is a too strong assumption that
is not valid in many cases. It restricts the wide use of the model. On the
other hand, this assumption allows us to apply the Cox model to solving the
explanation problems as a linear approximation of some unknown function of
covariates by considering coefficients of the covariates as quantitative
impacts on the prediction. As a result, Kovalev at al.
\cite{Kovalev-Utkin-Kasimov-20a} proposed an explanation method called
SurvLIME, which deals with censored data and can be regarded as an extension
of LIME on the case of survival data. The basic idea behind SurvLIME is to
apply the Cox model to approximate the black-box survival model at a local
area around a test example. SurvLIME used the quadratic norm to take into
account the distance between CHFs. Following ideas underlying SurvLIME, Utkin
et al. \cite{Utkin-Kovalev-Kasimov-20a} proposed a modification of SurvLIME
called SurvLIME-Inf. In contrast to SurvLIME, SurvLIME-Inf uses $L_{\infty}%
$-norm for defining distances between CHFs. SurvLIME-Inf significantly
simplifies the model and provides better results when a training set is small.
Another explanation model proposed by Kovalev and Utkin
\cite{Kovalev-Utkin-20} is called SurvLIME-KS. This model uses the well-known
Kolmogorov-Smirnov bounds to ensure robustness of the explanation model to
cases of a small amount of training data or outliers of survival data.

This paper presents a method for producing counterfactual explanations for
survival machine learning black-box models, which is based on the above
results concerning with explanation of these models. The approach allows us to
find a counterfactual whose SF is connected with the SF of the original
example by means of some conditions. One of the difficulties of solving the
counterfactual explanation problem is that the classes of examples are
implicitly defined through outcomes of a machine learning survival model in
the form of survival functions or cumulative hazard functions. Therefore, a
condition establishing the difference between mean times to event of the
original example and the counterfactual is proposed. For example, the mean
time to event of the counterfactual should be larger than the mean time to
event of the original example by the value $r$. The meaning of counterfactuals
in survival analysis under the above condition can be represented by the
statement: \textquotedblleft Your treatment was not successful because of a
small dose of the medicine (one tablet). If your dose had been three tablets,
the mean time of recession would have been increased till a required
value\textquotedblright. Here the difference between the required value of the
mean time to recession and the recent mean time to recession of the patient is
$r$. The algorithm implementing the method is called as SurvCoFact. It depends
on the black-box model used in a certain study. In particular, when the Cox
model is considered as a black-box model, the exact solution can be obtained
because the optimization problem for computing counterfactuals is reduced to a
standard convex programming. In a general case of the black-box model, the
optimization problem for computing counterfactuals is non-convex. Therefore,
one of the ways for solving the optimization problem is to use some heuristic
global optimization method. A heuristic global optimization method called
Particle Swarm Optimization (PSO), proposed by Eberhart and Kennedy
\cite{Kennedy-Eberhart-1995}, can be applied to solving the counterfactual
explanation problem. The method is a population-based stochastic optimization
technique based on swarm and motivated by the intelligent collective behavior
of some animals \cite{Wang-Tan-Liu-18}.

In summary, the following contributions are made in this paper:

\begin{enumerate}
\item A statement of the counterfactual explanation problem in the framework
of survival analysis is proposed and a criterion for defining counterfactuals
is introduced.

\item It is shown that the counterfactual explanation problem can be reduced
to a standard convex optimization problem with linear constraints when the
black-box model is the Cox model. This fact leads to an exact solution of the
counterfactual explanation problem.

\item The PSO algorithm is applied to solving the counterfactual explanation
problem in a general case when the black-box model may be arbitrary.

\item The proposed approaches are illustrated by means of numerical
experiments with synthetic and real data, which show the efficiency and
accuracy of the approaches.
\end{enumerate}

The paper is organized as follows. Related work is in Section 2. Basic
concepts of survival analysis and the Cox model are given in Section 3.
Section 4 contains the standard counterfactual explanation problem statement
and its extension on the case of survival analysis. In Section 5, it is shown
how to reduce the counterfactual explanation problem to the convex programming
by the black-box Cox model. The PSO algorithm is briefly introduced in Section
6. Its application to the counterfactual explanation problem is considered in
Section 7. Numerical experiments with synthetic data and real data are given
in Section 8. Concluding remarks can be found in Section 9.

\section{Related work}

\textbf{Local explanation methods. }Due to importance of the machine learning
model explanation in many applications, a lot of methods have been proposed to
explain black-box models locally. One of the first local explanation methods
is the Local Interpretable Model-agnostic Explanations (LIME)
\cite{Ribeiro-etal-2016}, which uses simple and easily understandable linear
models to approximate the predictions of black-box models locally. Following
the original LIME \cite{Ribeiro-etal-2016}, a lot of its modifications have
been developed due to a nice simple idea underlying the method to construct a
linear approximating model in a local area around a test example. These
modifications are ALIME \cite{Shankaranarayana-Runje-2019}, NormLIME
\cite{Ahern-etal-2019}, Anchor LIME \cite{Ribeiro-etal-2018}, LIME-Aleph
\cite{Rabold-etal-2019}, GraphLIME \cite{Huang-Yamada-etal-2020}, SurvLIME
\cite{Kovalev-Utkin-Kasimov-20a}, SurvLIME-KS \cite{Kovalev-Utkin-20}. Another
explanation method, which is based on the linear approximation, is the SHAP
\cite{Lundberg-Lee-2017,Strumbel-Kononenko-2010}, which takes a game-theoretic
approach for optimizing a regression loss function based on Shapley values. A
comprehensive analysis of LIME, including the study of its applicability to
different data types, for example, text and image data, is provided by Garreau
and Luxburg \cite{Garreau-Luxburg-2020}.

An important group of explanation methods is based on perturbation techniques
\cite{Fong-Vedaldi-2019,Fong-Vedaldi-2017,Petsiuk-etal-2018,Vu-etal-2019},
which are also used in LIME. The basic idea behind the perturbation techniques
is that contribution of a feature can be determined by measuring how a
prediction score changes when the feature is altered \cite{Du-Liu-Hu-2019}.
Perturbation techniques can be applied to a black-box model without any need
to access the internal structure of the model. However, the corresponding
methods are computationally complex when samples are of the high dimensionality.

A lot of explanation methods, their analysis, and critical review can be found
in survey papers
\cite{Adadi-Berrada-2018,Arrieta-etal-2019,Carvalho-etal-2019,Guidotti-2019,Rudin-2019,Xie-Ras-etal-2020}%
.

Most explanation methods deal with the point-valued results produced by
explainable black-box models, for example, with classes of examples. In
contrast to these models, outcomes of survival models are functions, for
example, SFs or CHFs. It follows that LIME should be extended on the case of
models with functional outcomes, in particular, with survival models. An
example of such extended models is SurvLIME \cite{Kovalev-Utkin-Kasimov-20a}.

\textbf{Counterfactual explanations}. In order to get intuitive and
human-friendly explanations, a lot of counterfactual explanation methods
\cite{Wachter-etal-2017} were developed by several authors
\cite{Buhrmester-etal-2019,Dandl-etal-2020,Fernandez-etal-20,Goyal-etal-2018,Guidotti-etal-2019a,Hendricks-etal-2018,Looveren-Klaise-2019,Lucic-etal-19,Poyiadzi-etal-2019,Russel-19,Vermeire-Martens-20,Waa-etal-2018,White-Garcez-2020}%
. The counterfactual explanation tells us what to do to achieve a desired outcome.

Some counterfactual explanation methods use combinations with other methods
like LIME and SHAP. For example, Ramon et al. \cite{Ramon-etal-2020} proposed
the so-called LIME-C and SHAP-C methods as counterfactual extensions of LIME
and SHAP, White and Garcez \cite{White-Garcez-2020} proposed the CLEAR methods
which can also be viewed as a combination of LIME and counterfactual explanations.

Many counterfactual explanation methods apply perturbation techniques to
examine feature perturbations that lead to a different outcome of a black-box
model. In fact, this is an approach to generate counterfactuals to alter an
input of the black-box model and to observe how the output changes. One of the
methods using perturbations is the Growing Spheres method
\cite{Laugel-etal-18}, which can be referred to counterfactual explanations to
some extent. The method determines the minimal changes needed to alter a
prediction. Perturbations are usually performed towards an interpretable
counterfactuals in a lot of methods
\cite{Dhurandhar-etal-2018,Dhurandhar-etal-2019,Looveren-Klaise-2019}.

Another direction for applying counterfactuals concerns with counterfactual
visual explanations which can be generated to explain the decisions of a deep
learning system by identifying what and how regions of an input image would
need to change in order for the system to produce a specified output
\cite{Goyal-etal-2018}. Hendricks et al. \cite{Hendricks-etal-18a} proposed
counterfactual explanations in natural language by generating counterfactual
textual evidence. Counterfactual \textquotedblleft feature-highlighting
explanations\textquotedblright\ were proposed by Barocas et al.
\cite{Barocas-etal-20} by highlighting a set of features deemed most relevant
and withholding others.

A lot of other approaches have been proposed in the last years, but there are
no counterfactual explanations of predictions provided by the survival machine
learning systems. Therefore, our aim of the presented work is to propose a new
method for counterfactual survival explanations.

\textbf{Machine learning models in survival analysis}. Survival analysis is an
important direction for taking into account censored data. It covers a lot of
real application problems, especially in medicine, reliability analysis, risk
analysis. Therefore, the machine learning approach for solving the survival
analysis problems allows improving the survival data processing. A
comprehensive review of the machine learning models dealing with survival
analysis problems was provided by Wang et al. \cite{Wang-Li-Reddy-2017}. One
of the most powerful and popular methods for dealing with survival data is the
Cox model \cite{Cox-1972}. A lot of its modifications have been developed in
order to relax some strong assumptions underlying the Cox model. In order to
take into account the high dimensionality of survival data and to solve the
feature selection problem with these data, Tibshirani \cite{Tibshirani-1997}
presented a modification based on the Lasso method. Similar Lasso
modifications, for example, the adaptive Lasso, were also proposed by several
authors \cite{Kim-etal-2012,Witten-Tibshirani-2010,Zhang-Lu-2007}. The next
extension of the Cox model is a set of SVM modifications
\cite{Khan-Zubek-2008,Widodo-Yang-2011}. Various architectures of neural
networks, starting from a simple network \cite{Faraggi-Simon-1995} proposed to
relax the linear relationship assumption in the Cox model, have been developed
\cite{Haarburger-etal-2018,Katzman-etal-2018,Ranganath-etal-2016,Zhu-Yao-Huang-2016}
to solve prediction problems in the framework of survival analysis. Despite
many powerful machine learning approaches for solving the survival problems,
the most efficient and popular tool for survival analysis under condition of
small survival data is the extension of the standard random forest
\cite{Breiman-2001} called the random survival forest (RSF)
\cite{Ibrahim-etal-2008,Mogensen-etal-2012,Wang-Zhou-2017,Wright-etal-2017}.

Most of the above models dealing with survival data can be regarded as
black-box models and should be explained. However, only the Cox model has a
simple explanation due to its linear relationship between covariates.
Therefore, it can be used to approximate more powerful models, including
survival deep neural networks and random survival forests, in order to explain
predictions of these models.

\section{Some elements of survival analysis}

\subsection{Basic concepts}

In survival analysis, an example (patient) $i$ is represented by a triplet
$(\mathbf{x}_{i},\delta_{i},T_{i})$, where $\mathbf{x}_{i}=(x_{i1}%
,...,x_{id})$ is the vector of the patient parameters (characteristics) or the
vector of the example features; $T_{i}$ is time to event of the example. If
the event of interest is observed, $T_{i}$ corresponds to the time between
baseline time and the time of event happening, in this case $\delta_{i}=1$,
and we have an uncensored observation. If the example event is not observed,
$T_{i}$ corresponds to the time between baseline time and end of the
observation, and the event indicator is $\delta_{i}=0$, and we have a censored
observation. Suppose a training set $D$ consists of $n$ triplets
$(\mathbf{x}_{i},\delta_{i},T_{i})$, $i=1,...,n$. The goal of survival
analysis is to estimate the time to the event of interest $T$ for a new
example (patient) with feature vector denoted by $\mathbf{x}$ by using the
training set $D$.

The survival and hazard functions are key concepts in survival analysis for
describing the distribution of event times. The SF denoted by $S(t|\mathbf{x}%
)$ as a function of time $t$ is the probability of surviving up to that time,
i.e., $S(t|\mathbf{x})=\Pr\{T>t|\mathbf{x}\}$. The hazard function
$h(t|\mathbf{x})$ is the rate of event at time $t$ given that no event
occurred before time $t$, i.e., $h(t|\mathbf{x})=f(t|\mathbf{x}%
)/S(t|\mathbf{x})$, where $f(t|\mathbf{x})$ is the density function of the
event of interest. The hazard rate is defined as
\begin{equation}
h(t|\mathbf{x})=-\frac{\mathrm{d}}{\mathrm{d}t}\ln S(t|\mathbf{x}).
\end{equation}

Another important concept is the CHF $H(t|\mathbf{x})$, which is defined as
the integral of the hazard function $h(t|\mathbf{x})$, i.e.,
\begin{equation}
H(t|\mathbf{x})=\int_{-\infty}^{t}h(x|\mathbf{x})dx.
\end{equation}

The SF can be expressed through the CHF as $S(t|\mathbf{x})=\exp\left(
-H(t|\mathbf{x})\right)  $.

\subsection{The Cox model}

Let us consider main concepts of the Cox proportional hazards model,
\cite{Hosmer-Lemeshow-May-2008}. According to the model, the hazard function
at time $t$ given predictor values $\mathbf{x}$ is defined as
\begin{equation}
h(t|\mathbf{x},\mathbf{b})=h_{0}(t)\Psi(\mathbf{x},\mathbf{b})=h_{0}%
(t)\exp\left(  \psi(\mathbf{x},\mathbf{b})\right)  .
\end{equation}

Here $h_{0}(t)$ is a baseline hazard function which does not depend on the
vector $\mathbf{x}$ and the vector $\mathbf{b}$; $\Psi(\mathbf{x})$ is the
covariate effect or the risk function; $\mathbf{b}=(b_{1},...,b_{d})$ is an
unknown vector of regression coefficients or parameters. It can be seen from
the above expression for the hazard function that the reparametrization
$\Psi(\mathbf{x},\mathbf{b})=\exp\left(  \psi(\mathbf{x},\mathbf{b})\right)  $
is used in the Cox model. The function $\psi(\mathbf{x},\mathbf{b})$ in the
model is linear, i.e.,
\begin{equation}
\psi(\mathbf{x},\mathbf{b})=\mathbf{x^{\mathrm{T}}b}=\sum\nolimits_{k=1}%
^{d}b_{k}x_{k}.
\end{equation}

In the framework of the Cox model, the SF $S(t|\mathbf{x},\mathbf{b})$ is
computed as
\begin{equation}
S(t|\mathbf{x},\mathbf{b})=\exp(-H_{0}(t)\exp\left(  \psi(\mathbf{x}%
,\mathbf{b})\right)  )=\left(  S_{0}(t)\right)  ^{\exp\left(  \psi
(\mathbf{x},\mathbf{b})\right)  }.
\end{equation}

Here $H_{0}(t)$ is the cumulative baseline hazard function; $S_{0}(t)$ is the
baseline SF. It is important to note that functions $H_{0}(t)$ and $S_{0}(t)$
do not depend on $\mathbf{x}$ and $\mathbf{b}$.

The partial likelihood in this case is defined as follows:
\begin{equation}
L(\mathbf{b})=\prod_{j=1}^{n}\left[  \frac{\exp(\psi(\mathbf{x}_{j}%
,\mathbf{b}))}{\sum_{i\in R_{j}}\exp(\psi(\mathbf{x}_{i},\mathbf{b}))}\right]
^{\delta_{j}}.
\end{equation}

Here $R_{j}$ is the set of patients who are at risk at time $t_{j}$. The term
\textquotedblleft at risk at time $t$\textquotedblright\ means patients who
die at time $t$ or later.

\section{Counterfactual explanation for survival models: problem statement}

We consider a definition of the counterfactual explanation proposed by Wachter
et al. \cite{Wachter-etal-2017} and rewrite it in terms of the survival models.

\begin{definition}
[\cite{Wachter-etal-2017}]Assume a prediction function $f$ is given. Computing
a counterfactual $\mathbf{z=}(z_{1},...,z_{d})\in\mathbb{R}^{d}$ for a given
input $\mathbf{x=}(x_{1},...,x_{d})\in\mathbb{R}^{d}$ is derived by solving
the following optimization problem:%
\begin{equation}
\min_{\mathbf{z}\in\mathbb{R}^{m}}\{l(f(\mathbf{z}),f(\mathbf{x}%
))+C\mu(\mathbf{z},\mathbf{x})\}, \label{Surv_counter_20}%
\end{equation}
where $l(\cdot,\cdot)$ denotes a loss function, which establishes a
relationship between the explainable black-box model outputs; $\mu(\cdot
,\cdot)$ a penalty term for deviations of $\mathbf{z}$ from the original input
$\mathbf{x}$, which is expressed through a distance between $\mathbf{z}$ and
$\mathbf{x}$, for example, the Euclidean distance; $C>0$ denotes the
regularization strength.
\end{definition}

The function $l(f(\mathbf{z}),f(\mathbf{x}))$ encourages the prediction of
$\mathbf{z}$ to be different in accordance with a certain rule than the
prediction of the original point $\mathbf{x}$. The penalty term $\mu
(\mathbf{z},\mathbf{x})$ minimizes the distance between $\mathbf{z}$ and
$\mathbf{x}$ with the aim to find nearest counterfactuals to $\mathbf{x}$. It
can be defined as
\begin{equation}
\mu(\mathbf{z},\mathbf{x})=\left\Vert \mathbf{z}-\mathbf{x}\right\Vert _{2}.
\label{Surv_counter_20_1}%
\end{equation}

It is important to note that the above optimization problem can be extended by
including additional terms. In particular, many algorithms of the
counterfactual explanation use a term which makes counterfactuals close to the
observed data. It can be done, for example, by minimizing the distance between
the counterfactual $\mathbf{z}$ and the $k$ nearest observed data points
\cite{Dandl-etal-2020} or by minimizing the distance between the
counterfactual $\mathbf{z}$ and the class prototypes
\cite{Looveren-Klaise-2019}.

Let us consider an analogy of survival models with the standard classification
models where all points are divided into classes. We also have to divide all
patients into classes by means of an implicit relationship between the
black-box survival model predictions. It is important to note that predictions
are the CHFs or the SFs. Therefore, the introduced loss function
$l(f(\mathbf{z}),f(\mathbf{x}))$ should take into account the difference
between the CHFs or the SFs to some extent, which characterize different
\textquotedblleft classes\textquotedblright\ or groups\ of patients. It is
necessary to establish the relationship between CHFs or between SFs, which
would separate groups of patients of interest. One of the simplest way is to
separate groups of patients in accordance with their feature vectors. However,
this can be done if the groups of patients are known, for example, the
treatment and control groups. In many cases, it is difficult to divide
patients into groups by relying on their features because this division does
not take into account outcomes, for example, SFs of patients.

Another way for separating patients is to consider the difference between the
corresponding mean times to events for counterfactual $\mathbf{z}$ and input
$\mathbf{x}$. Therefore, several conditions of counterfactuals taking into
account mean values can be proposed. The mean values can be defined as
follows:
\begin{equation}
m(\mathbf{z})=\mathbb{E}(\mathbf{z})=\int_{0}^{\infty}S(t|\mathbf{z}%
)\mathrm{d}t,\ m(\mathbf{x})=\mathbb{E}(\mathbf{x})=\int_{0}^{\infty
}S(t|\mathbf{x})\mathrm{d}t. \label{Surv_counter_21}%
\end{equation}

Then the optimization problem (\ref{Surv_counter_20}) can be rewritten as
follows:
\begin{equation}
\min_{\mathbf{z}\in\mathbb{R}^{d}}\left\{  l(m(\mathbf{z}),m(\mathbf{x}%
))+C\mu(\mathbf{z},\mathbf{x})\right\}  . \label{Surv_counter_22}%
\end{equation}

We suppose that a condition for a boundary of \textquotedblleft
classes\textquotedblright\ of patients\ can be defined by a predefined
smallest distance between mean values which is equal to $r$. In other words, a
counterfactual $\mathbf{z}$ is defined by the following condition:%
\begin{equation}
m(\mathbf{z})-m(\mathbf{x})\geq r. \label{Surv_counter_22_1}%
\end{equation}

The condition for \textquotedblleft classes\textquotedblright\ of patients can
be also written as
\begin{equation}
m(\mathbf{x})-m(\mathbf{z})\geq r. \label{Surv_counter_22_2}%
\end{equation}

Let us unite the above conditions by means of the function
\begin{equation}
\psi(\mathbf{z})=r-\theta(m(\mathbf{z})-m(\mathbf{x})).
\label{Surv_counter_22_3}%
\end{equation}
where the parameter $\theta\in\{-1,1\}$. In particular, condition
(\ref{Surv_counter_22_1}) corresponds to case $\theta=1$, condition
(\ref{Surv_counter_22_2}) corresponds to case $\theta=-1$.

It should be noted that several conditions of counterfactuals taking into
account mean values can be proposed here. We take the difference between the
mean time to events of explainable point $\mathbf{x}$ and the point
$\mathbf{z}$. For example, if it is known that a group of patients belong to a
certain disease, we can define a prototype $\mathbf{x}_{p}$ of the group of
patients and try to find the difference $m(\mathbf{z})-m(\mathbf{x}_{p})$.

Let us consider the so-called hinge-loss function
\begin{equation}
l(f(\mathbf{z}),f(\mathbf{x}))=\max\left(  0,r-(f(\mathbf{z})-f(\mathbf{x}%
))\right)  . \label{Surv_counter_24}%
\end{equation}

Its minimization encourages to increase the difference $f(\mathbf{z}%
)-f(\mathbf{x})$. Taking into account (\ref{Surv_counter_20_1}) and
(\ref{Surv_counter_22_3}), the entire loss function can be rewritten and the
following optimization problem is formulated:
\begin{equation}
\min_{\mathbf{z}\in\mathbb{R}^{m}}L(\mathbf{z})=\min_{\mathbf{z}\in
\mathbb{R}^{m}}\left\{  \max\{0,\psi(\mathbf{z})\}+C\left\Vert \mathbf{z}%
-\mathbf{x}\right\Vert _{2}\right\}  . \label{Surv_counter_24_1}%
\end{equation}

In sum, the optimization problem (\ref{Surv_counter_24_1}) has to be solved in
order to find counterfactuals $\mathbf{z}$. It can be regarded as the
constrained optimization problem:
\begin{equation}
\min_{\mathbf{z}\in\mathbb{R}^{m}}L(\mathbf{z})=\min_{\mathbf{z}\in
\mathbb{R}^{m}}\left\Vert \mathbf{z}-\mathbf{x}\right\Vert _{2},
\label{Surv_counter_24_2}%
\end{equation}
subject to
\begin{equation}
\psi(\mathbf{z})\leq0. \label{Surv_counter_24_3}%
\end{equation}

Generally, the function $\psi(\mathbf{z})$ is not convex and cannot be written
in an explicit form. This fact complicates the problem and restricts possible
methods for its solving. Therefore, we propose to use the well-known heuristic
method called the PSO. Let us return to the problem (\ref{Surv_counter_24_1})
by writing it in the similar form:
\begin{equation}
\min_{\mathbf{z}\in\mathbb{R}^{m}}L(\mathbf{z})=\min_{\mathbf{z}\in
\mathbb{R}^{m}}\left\{  \max\{0,C\psi(\mathbf{z})\}+\left\Vert \mathbf{z}%
-\mathbf{x}\right\Vert _{2}\right\}  . \label{Surv_counter_24_4}%
\end{equation}

This problem can be explained as follows. If point $\mathbf{z}$ is from the
feasible set defined by condition $\psi(\mathbf{z})\leq0$, then the choice of
$\mathbf{z}$ minimizes the distance between $\mathbf{z}$ and $\mathbf{x}$. If
point $\mathbf{z}$ does not belong to the feasible set ($\psi(\mathbf{z})>0$),
then the penalty $C\psi(\mathbf{z})$ is assigned. It is assumed that the value
of $C\psi(\mathbf{z})$ is much larger than $\left\Vert \mathbf{z}%
-\mathbf{x}\right\Vert _{2}$. Therefore, it is recommended to take large
values of $C$, for example, $C=10^{6}$.

\section{The exact solution for the Cox model\label{sec:Cox_exact}}

We prove below that problem (\ref{Surv_counter_24_4}) can be reduced to a
standard convex optimization problem with linear constraints and can be
exactly solved if the black-box model is the Cox model. In other words, the
counterfactual example $\mathbf{z}$ can be determined by solving a convex
optimization problem.

Let $t_{0}<t_{1}<...<t_{q}<t_{q+1}$ be the distinct times to event of interest
from the set $\{T_{1},...,T_{n}\}$, where $t_{0}=0$, $t_{1}=\min
_{k=1,...,n}T_{k}$ and $t_{q}=\max_{k=1,...,n}T_{k}$, $t_{q+1}=t_{q}%
+t_{\gamma}$, $t_{\gamma}$ is a parameter. We assume that there hold
$S(\tau|\mathbf{x})=1$ for $\tau=t_{0}$, $0<S(\tau|\mathbf{x})<1$ for
$\tau\geq t_{1}$, and $S(\tau|\mathbf{x})=0$ for $\tau>t_{q+1}$. Let
$\Omega=[t_{0},t_{q+1}]$ and divide it into $q+1$ subsets $\Omega
_{0},...,\Omega_{q}$ such that $\Omega_{q}=[t_{q},t_{q+1}]$, $\Omega
_{j}=[t_{j},t_{j+1})$, $\Omega=\cup_{j=0,...,q}\Omega_{j}$; $\Omega_{j}%
\cap\Omega_{k}=\emptyset$, $\forall j\neq k$. Introduce the indicator function
$\chi_{j}(t)$ taking the value $1$ if $t\in\Omega_{j}$, and $0$ otherwise.
Then the baseline SF $S_{0}(\tau)$ the SF $S(\tau|\mathbf{x})$ under condition
of using the Cox model can be represented as follows:
\begin{equation}
S_{0}(\tau)=\sum_{j=0}^{q}s_{0,j}\cdot\chi_{j}(\tau),\ s_{0,0}=1,
\end{equation}
and
\begin{equation}
S(\tau|\mathbf{x})=\sum_{j=0}^{q}s_{0,j}^{\exp\left(  \mathbf{z}^{\mathrm{T}%
}\mathbf{b}\right)  }\cdot\chi_{j}(\tau).
\end{equation}

Hence, the mean value is
\begin{align}
m(\mathbf{x})  &  =\int_{0}^{\infty}S(t|\mathbf{x})\mathrm{d}t=\int
_{0}^{\infty}\left[  \sum_{j=0}^{q}s_{0,j}^{\exp\left(  \mathbf{x}%
^{\mathrm{T}}\mathbf{b}\right)  }\chi_{j}(\tau)\right]  \mathrm{d}t\nonumber\\
&  =\sum_{j=0}^{q}s_{0,j}^{\exp\left(  \mathbf{x}^{\mathrm{T}}\mathbf{b}%
\right)  }\left[  \int_{0}^{\infty}\chi_{j}(t)\mathrm{d}t\right]  =\sum
_{j=0}^{q}\mu_{j}s_{0,j}^{\exp\left(  \mathbf{x}^{\mathrm{T}}\mathbf{b}%
\right)  },
\end{align}
where $\mu_{j}=t_{j+1}-t_{j}>0$.

Denote $u=\mathbf{z}^{\mathrm{T}}\mathbf{b}$ and consider the function
\begin{equation}
\pi(u)=\sum_{j=0}^{q}\mu_{j}s_{0,j}^{\exp\left(  u\right)  }.
\end{equation}

Compute the following limits:
\begin{equation}
\lim_{u\rightarrow-\infty}\pi(u)=\sum_{j=0}^{q}\mu_{j}\lim_{u\rightarrow
-\infty}s_{0,j}^{\exp\left(  u\right)  }=\sum_{j=0}^{q}\mu_{j}=t_{q+1}%
-t_{0}=t_{q+1},
\end{equation}%
\begin{equation}
\lim_{u\rightarrow\infty}\pi(u)=\sum_{j=0}^{q}\mu_{j}\lim_{u\rightarrow\infty
}s_{0,j}^{\exp\left(  u\right)  }=\mu_{0}=t_{1}-t_{0}=t_{1}.
\end{equation}

The derivative of $\pi(u)$ is
\begin{equation}
\frac{\mathrm{d}\pi(u)}{\mathrm{d}u}=\sum_{j=0}^{q}\mu_{j}\frac{\mathrm{d}%
}{\mathrm{d}u}\left[  s_{0,j}^{\exp\left(  u\right)  }\right]  =\sum_{j=0}%
^{q}\left[  \mu_{j}\ln(s_{0,j})\right]  \left(  s_{0,j}^{\exp\left(  u\right)
}\exp(u)\right)  .
\end{equation}

Note that $s_{0,j}^{\exp\left(  u\right)  }\exp(u)\geq0$ for all $j$ and $u$;
$\mu_{j}\ln(s_{0,j})\leq0$ for all $j$. Hence, there holds
\begin{equation}
\frac{\mathrm{d}\pi(u)}{\mathrm{d}u}\leq0,\ \forall u.
\end{equation}

The above means that the function $\pi(u)$ is non-increasing with $u$.
Moreover, it is positive because its limits are positive too. Let us consider
the function
\begin{equation}
\zeta(u)=r-\theta(\pi(u)-m(\mathbf{x})).
\end{equation}

It is obvious that there holds $m(\mathbf{x})\in\lbrack t_{1},t_{q}]$ for
arbitrary $\mathbf{x}$.

Let $\theta=1$. Then

\begin{enumerate}
\item $\zeta(u)$ is a non-decreasing monotone function;

\item $a_{+}=\lim_{u\rightarrow-\infty}\zeta(u)=r-t_{q+1}+m(\mathbf{x})$;

\item $b_{+}=\lim_{u\rightarrow+\infty}\zeta(u)=r-t_{1}+m(\mathbf{x})$;

\item $r\in(0,t_{q+1}-m(\mathbf{x})]$ (otherwise $\psi(\mathbf{z})$ will be
always positive).
\end{enumerate}

It follows from the above that $a_{+}\leq0<b_{+}$ and the function
$\zeta(u)=0$ at a single point $u_{+}$. By using numerical methods, we can
find point $u_{+}$. Since the set of solutions is defined by the inequality
$\psi(\mathbf{z})\leq0$, then it is equivalent to $\zeta(u)\leq0$ and $u\leq
u_{+}$ or $\mathbf{z}^{\mathrm{T}}\mathbf{b}-u_{+}\leq0$.

Let $\theta=-1$. Then

\begin{enumerate}
\item $\zeta(u)$ is a non-increasing monotone function;

\item $a_{-}=\lim_{u\rightarrow-\infty}\zeta(u)=r+t_{q+1}-m(\mathbf{x})$;

\item $b_{-}=\lim_{u\rightarrow+\infty}\zeta(u)=r+t_{1}-m(\mathbf{x})$;

\item $r\in(0,m(\mathbf{x})-t_{1}]$ (otherwise $\psi(\mathbf{z})$ will be
always positive).
\end{enumerate}

It follows from the above that $a_{-}<0\leq b_{-}$ and the function
$\zeta(u)=0$ at a single point $u_{-}$ which can be numerically computed.
Since the set of solutions is defined by the inequality $\psi(\mathbf{z}%
)\leq0$, then it is equivalent to $\zeta(u)\geq0$ and $u\geq u_{-}$ or
$-\mathbf{z}^{\mathrm{T}}\mathbf{b}+u_{-}\leq0$.

The above conditions for $r$ and the sets of solutions can be united
\begin{equation}
r\in\left(  0,\frac{1}{2}\left[  (1+\theta)\left(  t_{q+1}-m(\mathbf{x}%
)\right)  +(1-\theta)\left(  m(\mathbf{x})-t_{1}\right)  \right]  \right]  ,
\end{equation}%
\begin{equation}
\theta\mathbf{z}^{\mathrm{T}}\mathbf{b-}\frac{1}{2}\left[  (1+\theta
)u_{+}-(1-\theta)u_{-}\right]  \leq0. \label{Surv_counter_40}%
\end{equation}

Constraints (\ref{Surv_counter_24_3}) to the problem (\ref{Surv_counter_24_2}%
)-(\ref{Surv_counter_24_3}) become (\ref{Surv_counter_40}) which are linear
with $\mathbf{z}$. It can be seen from the objective function
(\ref{Surv_counter_24_2}) and constraints (\ref{Surv_counter_40}) that this
optimization problem is convex with linear constraints, It can be solved by
means of standard programming methods.

\section{Particle Swarm Optimization}

The PSO algorithm proposed by Eberhart and Kennedy
\cite{Kennedy-Eberhart-1995} can be viewed as a stochastic optimization
technique based on swarm. There are several survey papers devoted to the PSO
algorithms, for example, \cite{Wang-Tan-Liu-18,Wang-etal-15-PSO}. We briefly
introduce this algorithm below.

The PSO performs searching via a swarm of particles that updates from
iteration to iteration. In order to reach the optimal or suboptimal solution
to the optimization problem, each particle moves in the direction to its
previously best position (denoted as \textquotedblleft pbest\textquotedblright%
) and the global best position (denoted as \textquotedblleft
gbest\textquotedblright) in the swarm. Suppose that the function $f(u)$ where
$f:\mathbb{R}^{n}\rightarrow\mathbb{R}$ has to be minimized. The PSO is
implemented in the form of the following algorithm:

\begin{enumerate}
\item Initialization (zero iteration):

\begin{itemize}
\item $N$ particles $\left\{  u_{k}^{0}\right\}  _{k=1}^{N}$ and their
velocities $\left\{  v_{k}^{0}\right\}  _{k=1}^{N}$ are generated;

\item the best position $p_{k}^{0}=u_{k}^{0}$ of the particle $u_{k}^{0}$ is fixed;

\item the best solution $g^{0}=\arg\min_{k}f(p_{k}^{0})$ is fixed.
\end{itemize}

\item Iteration $t$ ($t=1,...,N_{\text{iter}}$):

\begin{itemize}
\item velocities are adjusted:
\begin{equation}
\left(  v_{k}^{t}\right)  _{i}=w\left(  v_{k}^{t-1}\right)  _{i}+r_{1}%
c_{1}\left(  p_{k}^{t-1}-u_{k}^{t-1}\right)  _{i}+r_{2}c_{2}\left(
g^{t-1}-u_{k}^{t-1}\right)  _{i},
\end{equation}

where $w$, $c_{1}$, $c_{2}$ are parameters, $r_{1}$ and $r_{2}$ are random
variables from the uniform distribution in interval $[0,1]$;

\item positions of particle are adjusted:
\begin{equation}
u_{k}^{t}=u_{k}^{t-1}+v_{k}^{t};
\end{equation}

\item the best positions of particle are adjusted:
\begin{equation}
p_{k}^{t}=\arg\min_{u\in P}f(u),\ P=\{p_{k}^{t-1},u_{k}^{t}\};
\end{equation}

\item the best solution is adjusted:
\begin{equation}
g^{t}=\arg\min_{k}f(p_{k}^{t}).
\end{equation}

\end{itemize}
\end{enumerate}

The problem has five parameters: the number of particles $N$; the number of
iterations $N_{\text{iter}}$; the inertia weight $w$; the coefficient for the
cognitive term $c_{1}$ (the cognitive term helps the particles for exploring
the search space); the coefficient for the social term $c_{2}$ (the social
term helps the particles for exploiting the search space).

It is clear that parameters $N$ as well as $N_{\text{iter}}$ should be as
large as possible. Its upper bound depends only on the time that we can spend
on iteration. We take $N=2000$, $N_{\text{iter}}=1000$.

Other parameters are selected by using some heuristics
\cite{Bai-10,Clerc-Kennedy-02}, namely,
\begin{equation}
w=\eta,\ c_{1}=\eta\phi_{1},\ c_{2}=\eta\phi_{2},
\end{equation}
where%
\begin{equation}
\eta=\frac{2\kappa}{\left\vert 2-\phi-\sqrt{\phi^{2}-4\phi}\right\vert
},\ \phi=\phi_{1}+\phi_{2}>4,\ \kappa\in\lbrack0,1].
\end{equation}

The following values of the above introduced parameters are often taken:
$\phi_{1}=\phi_{2}=2.05$, $\kappa=1$. Hence, there hold $w=0.729$,
$c_{1}=c_{2}=1.4945$.

Particles are generated by means of the uniform distributions $U$ with the
following parameters:
\begin{equation}
\left(  u_{k}^{0}\right)  _{i}\sim U\left(  u_{i}^{\min},u_{i}^{\max}\right)
,\ u_{i}^{\min},u_{i}^{\max}\in\mathbb{R}.
\end{equation}

Velocities are similarly generated as:%
\begin{equation}
\left(  v_{k}^{0}\right)  _{i}\sim U\left(  -\left\vert u_{i}^{\max}%
-u_{i}^{\min}\right\vert ,-\left\vert u_{i}^{\max}-u_{i}^{\min}\right\vert
\right)  .
\end{equation}

\section{Application of the PSO to the survival counterfactual explanation}

Let us return to the counterfactual explanation problem in the framework of
survival analysis. Suppose that there exists a dataset $D$ with triplets
$(\mathbf{x}_{j},\delta_{j},T_{j})$, where $\mathbf{x}_{j}\in\mathbb{R}^{d}$,
$T_{j}>0$, $\delta_{j}\in\{0,1\}$. It is assumed that the explained machine
learning model $q(\mathbf{x})$ is trained on $D$. It should be noted that the
prediction of the machine learning survival model is the SF or the CHF, which
can be used for computing the mean time to event of interest $m(\mathbf{x})$.
In order to find the counterfactual $\mathbf{z}$, we have to solve the
optimization problem (\ref{Surv_counter_24_4}) with fixed $\mathbf{x}$, $r$,
and $C$.

Let us calculate bounds of the domain $\mathbf{x}$ for every feature on the
basis of the training set as
\begin{equation}
x_{i}^{\min}=\min_{j}\left\{  \left(  x_{j}\right)  _{i}\right\}
,\ x_{i}^{\max}=\max_{j}\left\{  \left(  x_{j}\right)  _{i}\right\}  .
\end{equation}

According to the PSO algorithm, the initial positions of particles are
generated as
\begin{equation}
\left(  u_{k}^{0}\right)  _{i}\sim U\left(  x_{i}^{\min},x_{i}^{\max}\right)
.
\end{equation}

So, the optimal solution can be found in the hyperparallelepiped $\mathcal{X}%
$:
\begin{equation}
\mathcal{X}=\left[  x_{1}^{\min},x_{1}^{\max}\right]  \times...\times\left[
x_{d}^{\min},x_{d}^{\max}\right]  .
\end{equation}

If there exists at least one point $\mathbf{x}_{j}^{\ast}$ in the training set
such that $\psi(\mathbf{x}_{j}^{\ast})\leq0$, then the region $\mathcal{X}$
can be adjusted. Let
\begin{equation}
\mathbf{z}_{closest,train}=\mathbf{z}_{ct}=\arg\min_{j}L(\mathbf{x}_{j}).
\end{equation}

Let us introduce a sphere $\mathcal{B}=\mathbb{B}(\mathbf{x},R_{ct})$ with
center $\mathbf{x}$ and radius $R_{ct}=\left\Vert \mathbf{x}-\mathbf{z}%
_{ct}\right\Vert _{2}$. The sphere can be partially located inside the
hyperparallelepiped $\mathcal{X}$ or can be larger it. Therefore, we restrict
the set of solutions by a set $\mathcal{M}$ defined as $\mathcal{M}%
=\mathcal{X}\cap\mathcal{B}$. To disable a possible passage beyond the limits
of $\mathcal{M}$, we introduce the restriction procedure, denoted as $RP$,
which supports that:

\begin{enumerate}
\item $\mathbf{z}=\mathbf{x}+\min\left\{  \left\Vert \mathbf{x}-\mathbf{z}%
\right\Vert _{2},R_{ct}\right\}  \frac{\mathbf{z}-\mathbf{x}}{\left\Vert
\mathbf{x}-\mathbf{z}\right\Vert _{2}}$

\item Loop: over all features of $\mathbf{z}$:

\begin{enumerate}
\item $\left(  \mathbf{z}\right)  _{i}=\min\left\{  \left(  \mathbf{z}\right)
_{i},x_{i}^{\max}\right\}  $

\item $\left(  \mathbf{z}\right)  _{i}=\max\left\{  \left(  \mathbf{z}\right)
_{i},x_{i}^{\min}\right\}  $
\end{enumerate}
\end{enumerate}

The initial positions of particles are generated as follows:
\begin{equation}
u_{1}^{0}=\mathbf{z}_{ct},\ u_{2}^{0},...,u_{N}^{0}\sim U(\mathcal{B}),
\end{equation}
and the above restriction procedure is used for all points except the first
one: $u_{k}^{0}=RP(u_{k}^{0})$, $k=2,...,N$. Positions of particle will be
adjusted by using the following expression:
\begin{equation}
u_{k}^{t}=RP\left(  u_{k}^{t-1}+v_{k}^{t}\right)  .
\end{equation}

Initial values of velocities are taken as $v_{k}^{0}=0$, $k=1,...,N$.

Let us point out properties of the above approach:

\begin{itemize}
\item The optimization results are always located in the set $\mathcal{M}$.

\item The \textquotedblleft worst\textquotedblright\ optimal solution is
$\mathbf{z}_{ct}$ because the optimization algorithm remembers the point
$\mathbf{z}_{ct}$ at the zero iteration as optimal, and the next iterations
never give the worse solution, if every initial position $u_{k}^{0}$ starting
from $k=2$ is out of the feasible set of solutions, i.e., $\psi(u_{k}^{0})>0$.
\end{itemize}

\section{Numerical experiments}

To perform numerical experiments, we use the following general scheme.

1. The Cox model and the RSF are considered as black-box models that are
trained on synthetic or real survival data. The outputs of the trained models
in the testing phase are SFs.

2. In order to study the proposed explanation algorithm by means of synthetic
data, we generate random survival times to events by using the Cox model estimates.

In order to analyze the numerical results, the following schemes of their
verification is proposed. When the Cox model is used as a black-box model, we
can get exact solution (see Sec. \ref{sec:Cox_exact}). This implies that we
can exactly compute the counterfactual $\mathbf{z}_{ver}$. Adding the
condition that the solution belongs to the hyperparallelepiped $\mathcal{X}$
to the problem with objective function (\ref{Surv_counter_24_2}) and
constraints (\ref{Surv_counter_40}), we use this solution ($\mathbf{z}_{ver}$)
as a referenced solution in order to compare another solution ($\mathbf{z}%
_{opt}$) obtained by means of the PSO. The Euclidean distance between
$\mathbf{z}_{ver}$ and $\mathbf{z}_{opt}$ can be a measure for the PSO
algorithm accuracy in the case of the black-box Cox model.

The next question is how to verify results of the RSF as a black-box model.
The problem is that the RSF does not allow us to obtain exact results by means
of formal methods, for example, by solving the optimization problem
(\ref{Surv_counter_24_2}). However, the counterfactual can be found with
arbitrary accuracy by considering all points or many points in accordance with
a grid. Then the minimal distance between the original point $\mathbf{x}$ and
each generated point is minimized under condition $\psi(\mathbf{z})\leq0$
which is verified for every generated $\mathbf{z}$. This is a computationally
extremely complex task, but it can be applied to testing results. By using the
above approach, many random points are generated from the set $\mathcal{M}$
defined in the previous section and approximate the optimum $\mathbf{z}_{ver}%
$. Random points for verification of results obtained by using the RSF are
uniformly selected from sphere $\mathcal{B}$ by using the restriction
procedure $RP$. The number of the points is taken $10^{6}$. In fact, this
approach can be regarded as the perturbation method with the exhaustive search.

When the black-box model is the RSF, then, in contrast to the black-box Cox
model, the accuracy of the proposed method is defined by the relationship
between two distances $\left\Vert \mathbf{z}_{ver}-\mathbf{x}\right\Vert _{2}$
and $\left\Vert \mathbf{z}_{opt}-\mathbf{x}\right\Vert _{2}$, which will be
denoted as $A$. The large values of $A$ indicate that the PSO provides better
results than the procedure of computing $\mathbf{z}_{ver}$ by means of many
generated points.

\subsection{Numerical experiments with synthetic data}

\subsubsection{Initial parameters of numerical experiments with synthetic
data}

Random survival times to events are generated by using the Cox model
estimates. An algorithm proposed by Bender et al. \cite{Bender-etal-2005} for
survival time data for the Cox model with Weibull distributed survival times
is applied to generate the random times. The Weibull distribution for
generation has the scale $\lambda_{0}=10^{-5}$ and shape $v=2$ parameters. For
experiments, we generate two types of data having the dimension $2$ and $20$,
respectively. The two-dimensional feature vectors are used in order to
graphically illustrate results of numerical experiments. The corresponding
feature vectors $\mathbf{x}$ are uniformly generated from hypercubes
$[0,1]^{2}$ and $[0,1]^{20}$. Random survival times $T_{j}$, $j=1,...,N$, are
generated in accordance with \cite{Bender-etal-2005} by using parameters
$\lambda_{0}$, $v$, $\mathbf{b}$ as follows:%
\begin{equation}
T_{j}=\left(  \frac{-\ln(\xi_{j})}{\lambda_{0}\exp\left(  \mathbf{x_{j}%
^{\mathrm{T}}b}\right)  }\right)  ^{1/v}, \label{KS_Inf_SurvLIME_84}%
\end{equation}
where $\xi_{j}$ is the $j$-th random variable uniformly distributed in
interval $[0,1]$; vectors of coefficients $\mathbf{b}$ are randomly selected
from hypercubes $[0,1]^{2}$ and $[0,1]^{20}$.

The event indicator $\delta_{j}$ is generated from the binomial distribution
with probabilities $\Pr\{\delta_{j}=1\}=0.9$, $\Pr\{\delta_{j}=0\}=0.1$.

For testing, two points are randomly selected from the hyperparallelepiped
$\mathcal{X}$ in accordance with two cases: $\theta=1$, condition
(\ref{Surv_counter_22_1}), and $\theta=-1$, condition (\ref{Surv_counter_22_2}%
). For each point, two tasks are solved: with parameter $\theta=1$ and
parameter $\theta=-1$. Parameter $r$ is also selected randomly for every task.

\subsubsection{The black-box Cox model}

\begin{figure}
[ptb]
\begin{center}
\includegraphics[
height=2.5469in,
width=4.7435in
]%
{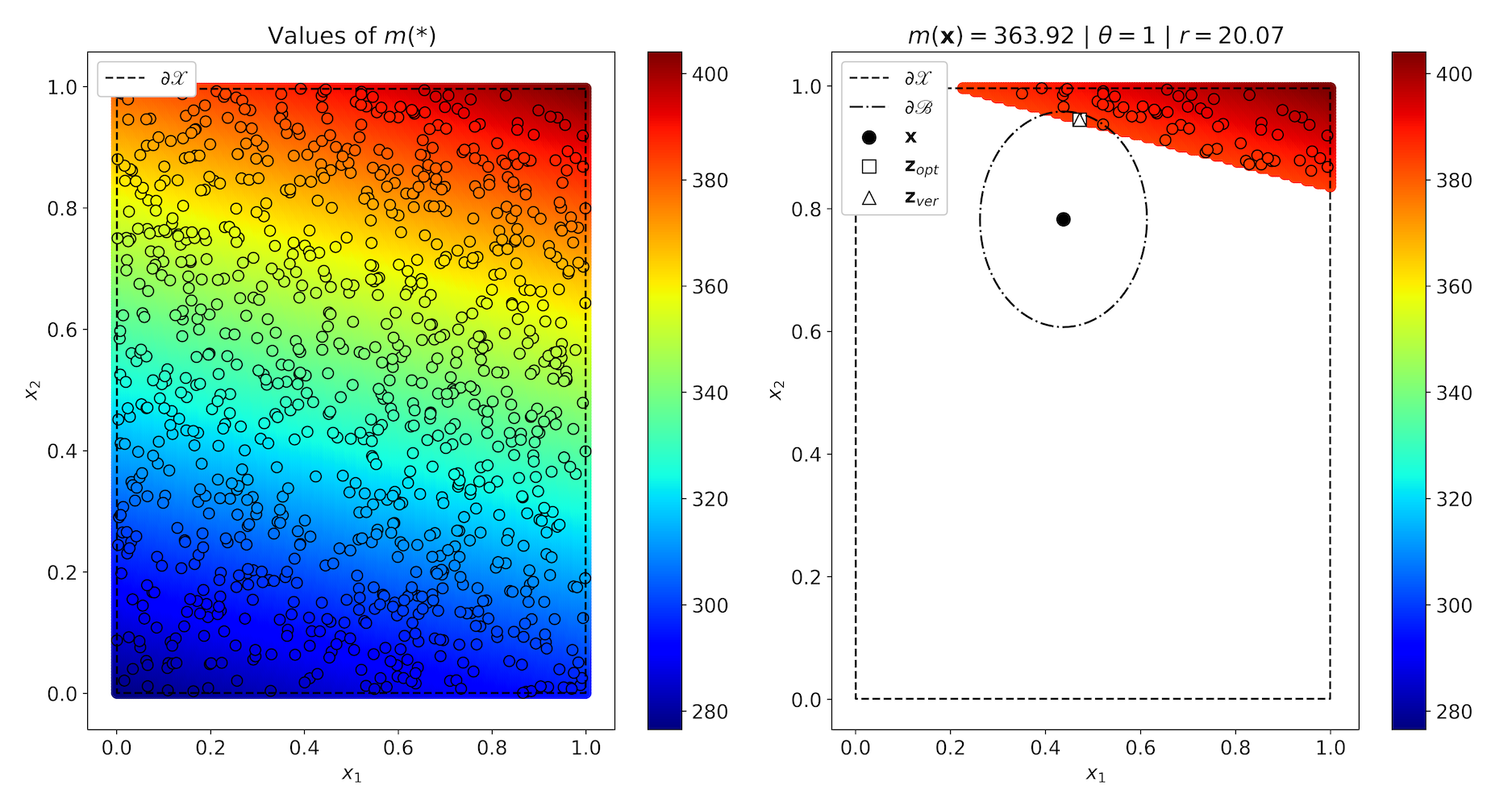}%
\caption{Original and counterfactual points by $\theta=1$ for the black-box
Cox model trained on synthetic data}%
\label{f:syn_cox_1}%
\end{center}
\end{figure}

\begin{figure}
[ptb]
\begin{center}
\includegraphics[
height=2.5555in,
width=4.7617in
]%
{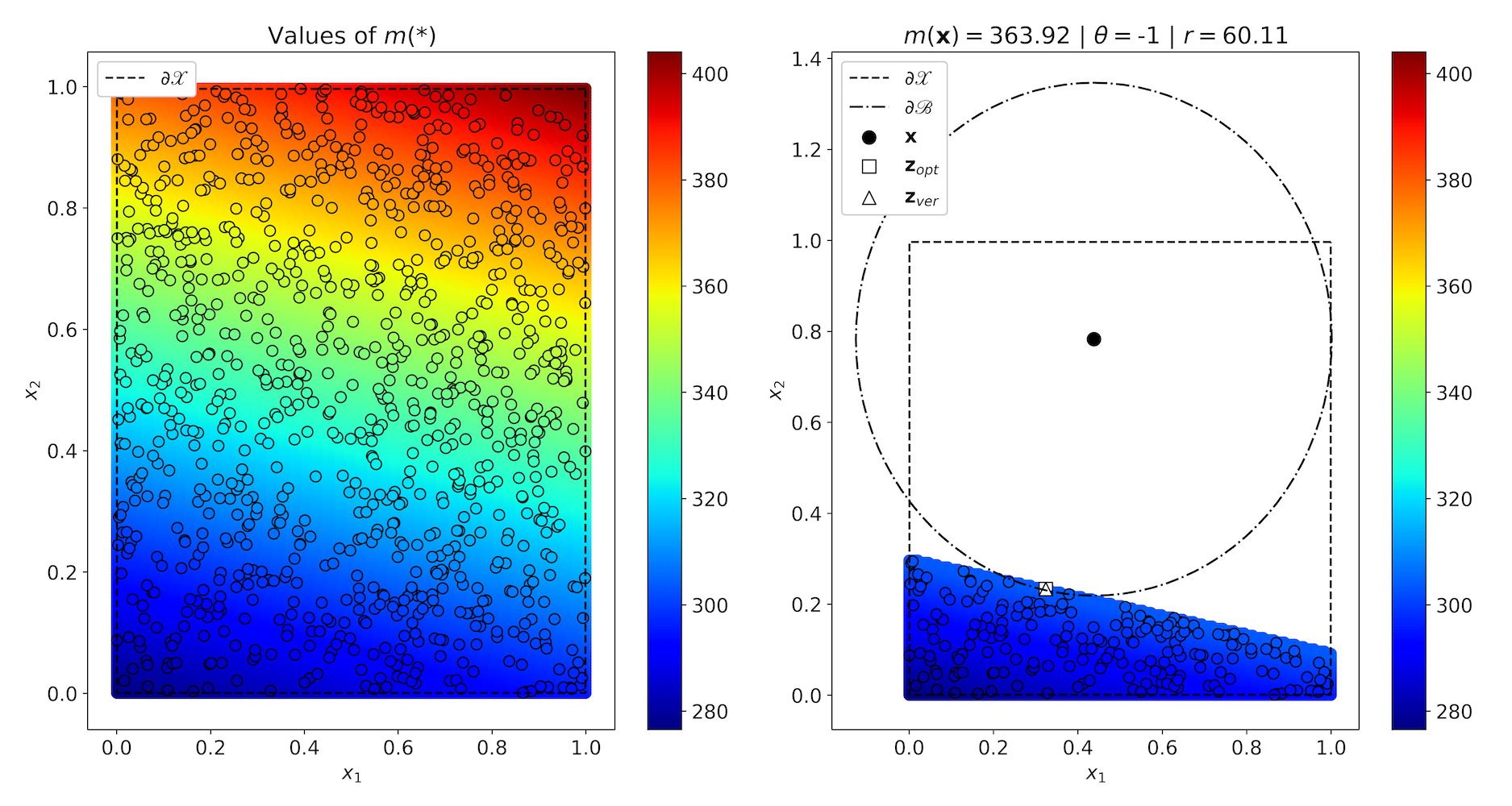}%
\caption{Original and counterfactual points by $\theta=-1$ for the black-box
Cox model trained on synthetic data}%
\label{f:syn_cox_2}%
\end{center}
\end{figure}

The first part of numerical experiments is performed with the black-box Cox
model and aims to show how results obtained by means of the PSO approximate
the verified results obtained as the solution of the convex optimization
problem with objective function (\ref{Surv_counter_24_2}) and constraints
(\ref{Surv_counter_40}). These results are illustrated in Figs.
\ref{f:syn_cox_1}-\ref{f:syn_cox_2}. The left picture in Fig.
\ref{f:syn_cox_1} shows how $m(\mathbf{x})$ changes depending on values of two
features $x_{1}$ and $x_{2}$. It can be seen from the picture that
$m(\mathbf{x})$ takes values from $280$ (the bottom left corner) till $400$
(the top right corner). Values of $m(\mathbf{x})$ are represented by means of
color. Small circles in the picture correspond to training examples. The bound
for the hyperparallelepiped $\mathcal{X}$ is denoted as $\partial\mathcal{X}$
and depicted in Fig. \ref{f:syn_cox_1} by the dashed line. The right picture
in Fig. \ref{f:syn_cox_1} displays the results of solving the problem for the
case $\theta=1$. The light background is the region outside the feasible
region defined by condition $\psi(\mathbf{z})\leq0$. The filled area
corresponds to condition $\psi(\mathbf{z})\leq0$. The bound for the sphere
$\mathcal{B}$ is denoted as $\partial\mathcal{B}$ and depicted in Fig.
\ref{f:syn_cox_1} by the dash-dot line. The explained point $\mathbf{x}$, the
verified solution $\mathbf{z}_{ver}$, and the solution obtained by the PSO
$\mathbf{z}_{opt}$ are depicted in Fig. \ref{f:syn_cox_1} by the black circle,
square, and triangle, respectively. Parameters of the corresponding numerical
experiment, including $m(\mathbf{x})$, $\theta$, $r$, are presented above the
right picture. It can be seen from Fig. \ref{f:syn_cox_1} that points
$\mathbf{z}_{ver}$ and $\mathbf{z}_{opt}$ are almost coincide. The same
results are illustrated in Fig. \ref{f:syn_cox_2} for the case $\theta=-1$.%

%

Similar results cannot be shown for the second type of synthetic data when
feature vectors have the dimensionality $20$. Therefore, we give them in Table
\ref{t:Contraf_Syn_Cox} jointly with numerical results for the two-dimensional
data. Parameters $r_{ver}$ and $r_{opt}$ in Table \ref{t:Contraf_Syn_Cox} are
defined as
\begin{equation}
r_{ver}=\theta\left(  m(\mathbf{z}_{ver})-m(\mathbf{x})\right)  ,
\end{equation}%
\begin{equation}
r_{opt}=\theta\left(  m(\mathbf{z}_{opt})-m(\mathbf{x})\right)  ,
\end{equation}
respectively. They show the relationship between original margin $r$ and
margins $r_{ver}$ and $r_{opt}$ obtained by means of the proposed methods. In
fact, values of $r_{ver}$ and $r_{opt}$ indicate how the obtained
counterfactuals fulfil condition (\ref{Surv_counter_22_1}) or condition
(\ref{Surv_counter_22_2}), i.e., conditions
\begin{equation}
\psi(\mathbf{z}_{ver})=r-r_{ver}\leq0,\ \psi(\mathbf{z}_{opt})=r-r_{opt}\leq0.
\end{equation}

The last three columns also display the relationship between $\mathbf{z}%
_{ver}$, $\mathbf{z}_{opt}$ and $\mathbf{x}$. In particular, the value of
$\left\Vert \mathbf{z}_{ver}-\mathbf{z}_{opt}\right\Vert _{2}$ can be regarded
as the accuracy measure of the obtained counterfactual.%

\begin{table}[tbp] \centering
\caption{Results of numerical experiments for the black-box Cox model trained on synthetic data}%
\begin{tabular}
[c]{cccccccc}\hline
$d$ & $\theta$ & $r$ & $r_{ver}$ & $r_{opt}$ & $\left\Vert \mathbf{z}%
_{ver}-\mathbf{x}\right\Vert _{2}$ & $\left\Vert \mathbf{z}_{opt}%
-\mathbf{x}\right\Vert _{2}$ & $\left\Vert \mathbf{z}_{ver}-\mathbf{z}%
_{opt}\right\Vert _{2}$\\\hline
& $1$ & $42.80$ & $42.80$ & $42.80$ & $0.367$ & $0.367$ & $4.76\times10^{-6}%
$\\\cline{2-8}%
$2$ & $-1$ & $40.50$ & $40.50$ & $40.50$ & $0.395$ & $0.395$ & $1.01\times
10^{-6}$\\\cline{2-8}
& $1$ & $20.07$ & $20.07$ & $20.07$ & $0.166$ & $0.166$ & $3.72\times10^{-7}%
$\\\cline{2-8}
& $-1$ & $60.11$ & $60.11$ & $60.11$ & $0.561$ & $0.561$ & $9.93\times10^{-8}%
$\\\hline
& $1$ & $238.94$ & $238.94$ & $238.94$ & $0.322$ & $0.322$ & $1.39\times
10^{-2}$\\\cline{2-8}%
$20$ & $-1$ & $206.29$ & $206.29$ & $206.29$ & $0.476$ & $0.476$ &
$1.34\times10^{-2}$\\\cline{2-8}
& $1$ & $315.33$ & $315.33$ & $315.33$ & $0.461$ & $0.461$ & $7.86\times
10^{-3}$\\\cline{2-8}
& $-1$ & $91.86$ & $91.86$ & $91.86$ & $0.204$ & $0.205$ & $1.99\times10^{-2}%
$\\\hline
\end{tabular}
\label{t:Contraf_Syn_Cox}%
\end{table}%

\subsubsection{The black-box RSF}

\begin{figure}
[ptb]
\begin{center}
\includegraphics[
height=2.4146in,
width=4.5005in
]%
{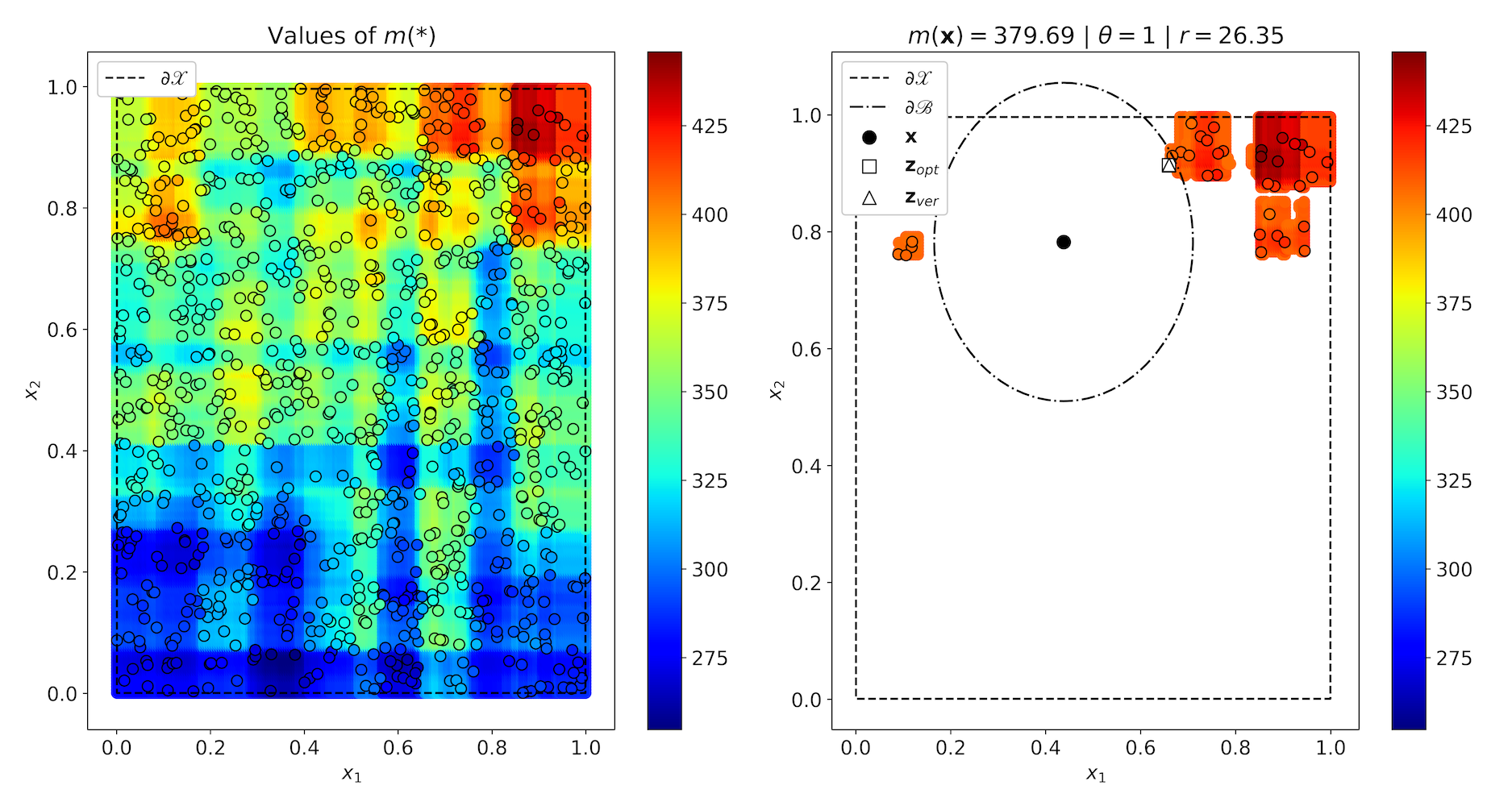}%
\caption{Original and counterfactual points by $\theta=1$ for the black-box
RSF trained on synthetic data}%
\label{f:syn_rsf_1}%
\end{center}
\end{figure}

\begin{figure}
[ptb]
\begin{center}
\includegraphics[
height=2.4293in,
width=4.5273in
]%
{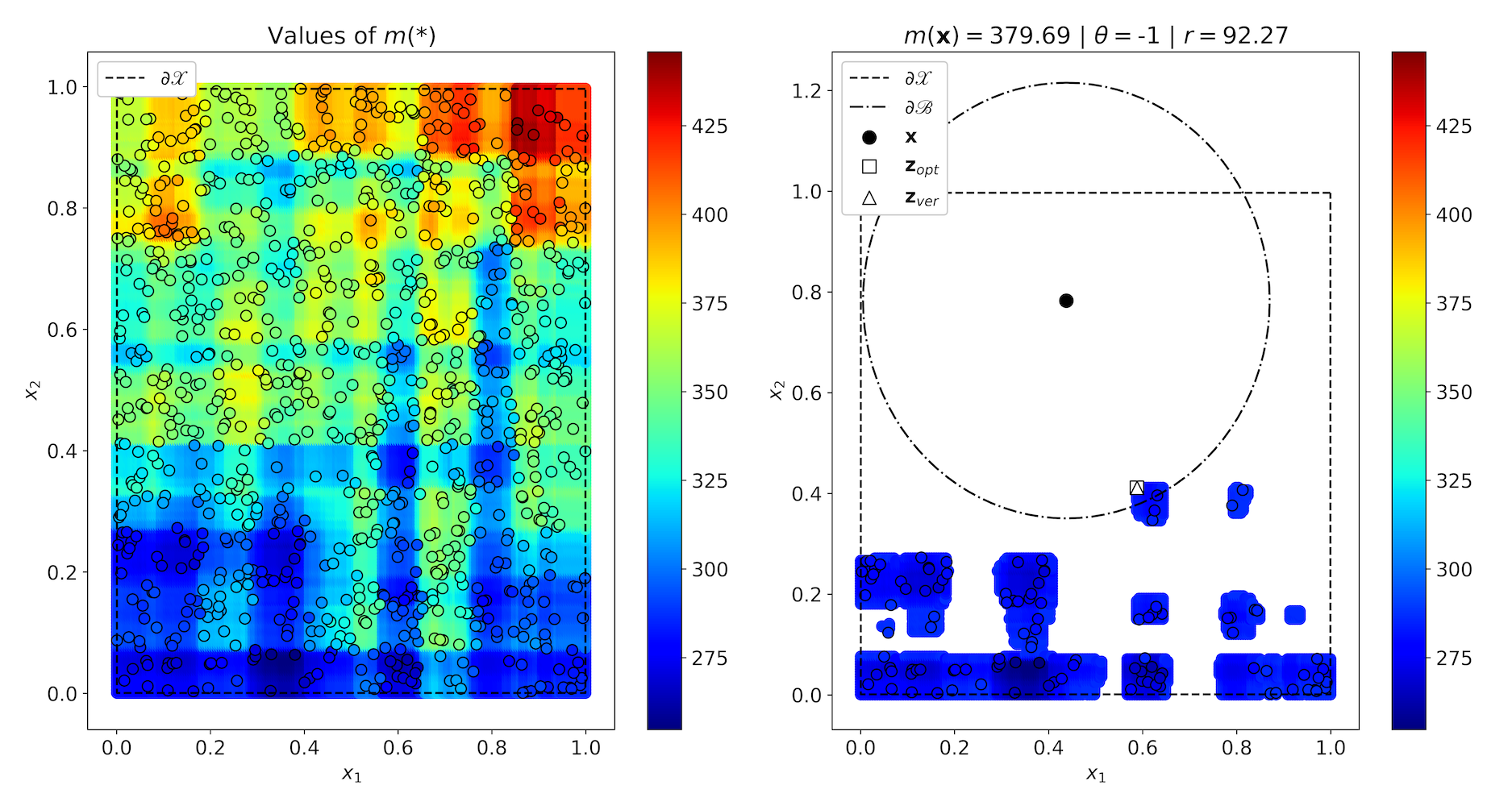}%
\caption{Original and counterfactual points by $\theta=-1$ for the black-box
RSF trained on synthetic data}%
\label{f:syn_rsf_2}%
\end{center}
\end{figure}

The second part of numerical experiments is performed with the RSF as a
black-box model. The RSF consists of $250$ decision survival trees. The
results are shown in Figs. \ref{f:syn_rsf_1}-\ref{f:syn_rsf_2}. In this cases,
$\mathbf{z}_{ver}$ is computed by generating many points ($10^{6}$) and
computing $m(\mathbf{z})$ for each point. The counterfactual $\mathbf{z}%
_{ver}$ minimizes the distance $\left\Vert \mathbf{z}_{ver}-\mathbf{x}%
\right\Vert $ under condition $\psi(\mathbf{z}_{ver})\leq0$. It can be seen
from the pictures that $\mathbf{z}_{ver}$ is again very close to
$\mathbf{z}_{opt}$.%

%

Results of experiments with training data having two- and twenty-dimensional
feature vectors are given Table \ref{t:Contraf_Syn_RSF}. It can be seen from
Table \ref{t:Contraf_Syn_RSF} that $\mathbf{z}_{opt}$ is not worse ($d=2$) or
even better ($d=20$) than $\mathbf{z}_{ver}$ obtained by means of generating
the large number of random points (see velues of $A$).%

\begin{table}[tbp] \centering
\caption{Results of numerical experiments for the black-box RSF trained on synthetic data}%
\begin{tabular}
[c]{cccccccc}\hline
$d$ & $\theta$ & $r$ & $r_{ver}$ & $r_{opt}$ & $\left\Vert \mathbf{z}%
_{ver}-\mathbf{x}\right\Vert _{2}$ & $\left\Vert \mathbf{z}_{opt}%
-\mathbf{x}\right\Vert _{2}$ & $A$\\\hline
& $1$ & $74.81$ & $75.67$ & $75.39$ & $0.153$ & $0.153$ & $0.0007$%
\\\cline{2-8}%
$2$ & $-1$ & $46.91$ & $49.32$ & $47.59$ & $0.346$ & $0.346$ & $0.0003$%
\\\cline{2-8}
& $1$ & $26.35$ & $27.93$ & $26.79$ & $0.259$ & $0.258$ & $0.0008$%
\\\cline{2-8}
& $-1$ & $92.27$ & $92.31$ & $92.30$ & $0.401$ & $0.401$ & $0.0006$\\\hline
& $1$ & $229.55$ & $232.39$ & $229.72$ & $0.969$ & $0.763$ & $0.2062$%
\\\cline{2-8}%
$20$ & $-1$ & $133.46$ & $135.84$ & $133.50$ & $0.876$ & $0.563$ &
$0.3136$\\\cline{2-8}
& $1$ & $249.88$ & $265.91$ & $250.15$ & $0.982$ & $0.641$ & $0.3414$%
\\\cline{2-8}
& $-1$ & $63.30$ & $63.56$ & $63.40$ & $0.570$ & $0.220$ & $0.3495$\\\hline
\end{tabular}
\label{t:Contraf_Syn_RSF}%
\end{table}%

\subsection{Numerical experiments with real data}

We consider the following real datasets to study the proposed approach:
Stanford2 and Myeloid. The datasets can be downloaded via R package
\textquotedblleft survival\textquotedblright.

The dataset Stanford2 consists of survival data of patients on the waiting
list for the Stanford heart transplant program. It contains 184 patients. The
number of features is 2 plus 3 variables: time to death, the event indicator,
the subject identifier.

The dataset Myeloid is based on a trial in acute myeloid leukemia. It contains
646 patients. The number of features is 5 plus 3 variables: time to death, the
event indicator, the subject identifier. In this dataset, we do not consider
the feature \textquotedblleft sex\textquotedblright\ because it cannot be
changed. Moreover, we consider two cases for the feature \textquotedblleft
trt\textquotedblright\ (treatment arm), when it takes values \textquotedblleft
A\textquotedblright\ and \textquotedblleft B\textquotedblright. In other
words, we divide all patients into two groups depending on the treatment arm.
As a result, we have three datasets: Stanford2 and Myeloid-A and Myeloid-B.

\subsubsection{The black-box Cox model}

\begin{figure}
[ptb]
\begin{center}
\includegraphics[
height=2.5322in,
width=4.7176in
]%
{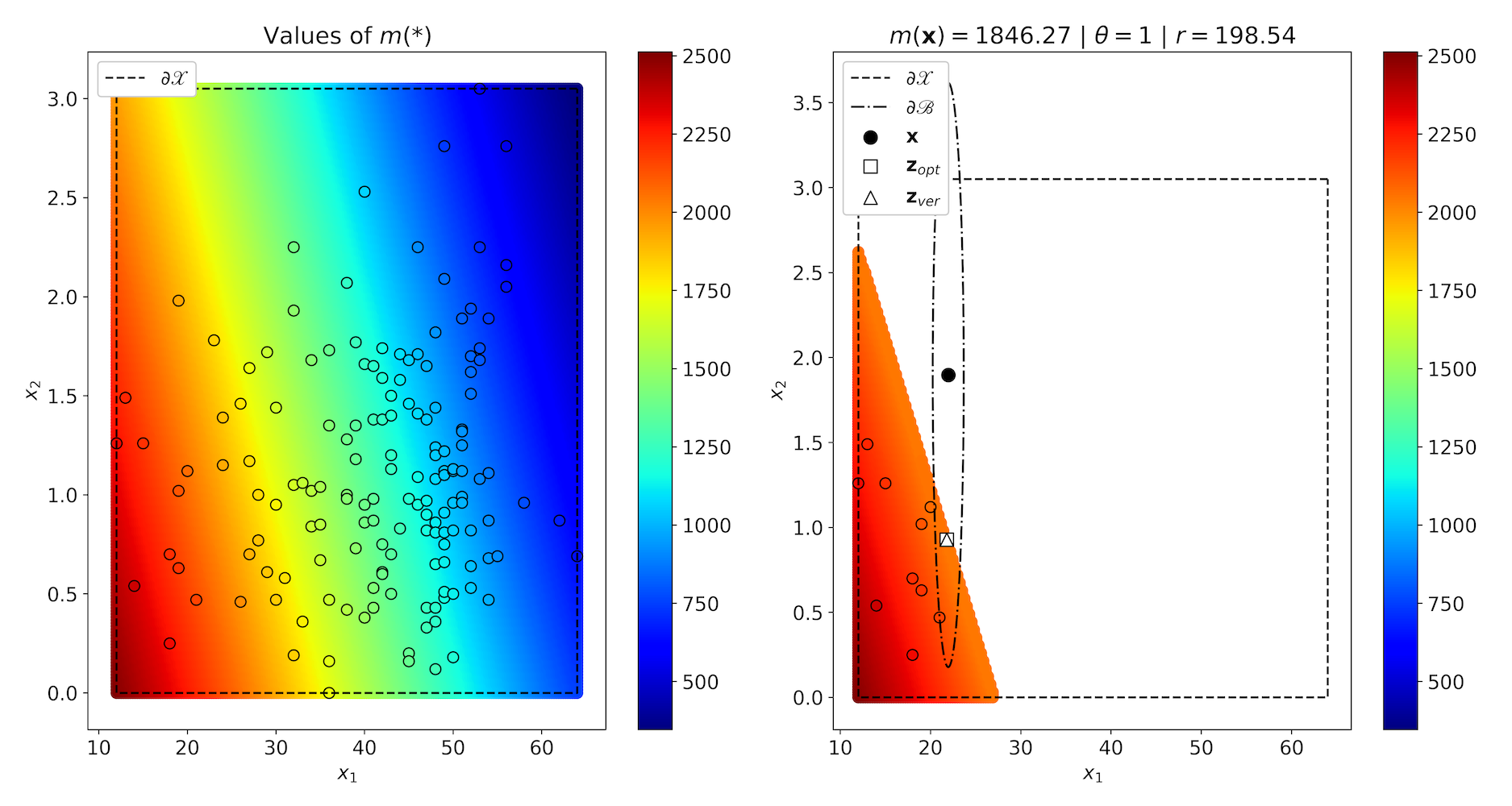}%
\caption{Original and counterfactual points by $\theta=1$ for the black-box
Cox model trained on the dataset Stanford2}%
\label{stanford2_cox_1}%
\end{center}
\end{figure}

\begin{figure}
[ptb]
\begin{center}
\includegraphics[
height=2.5512in,
width=4.753in
]%
{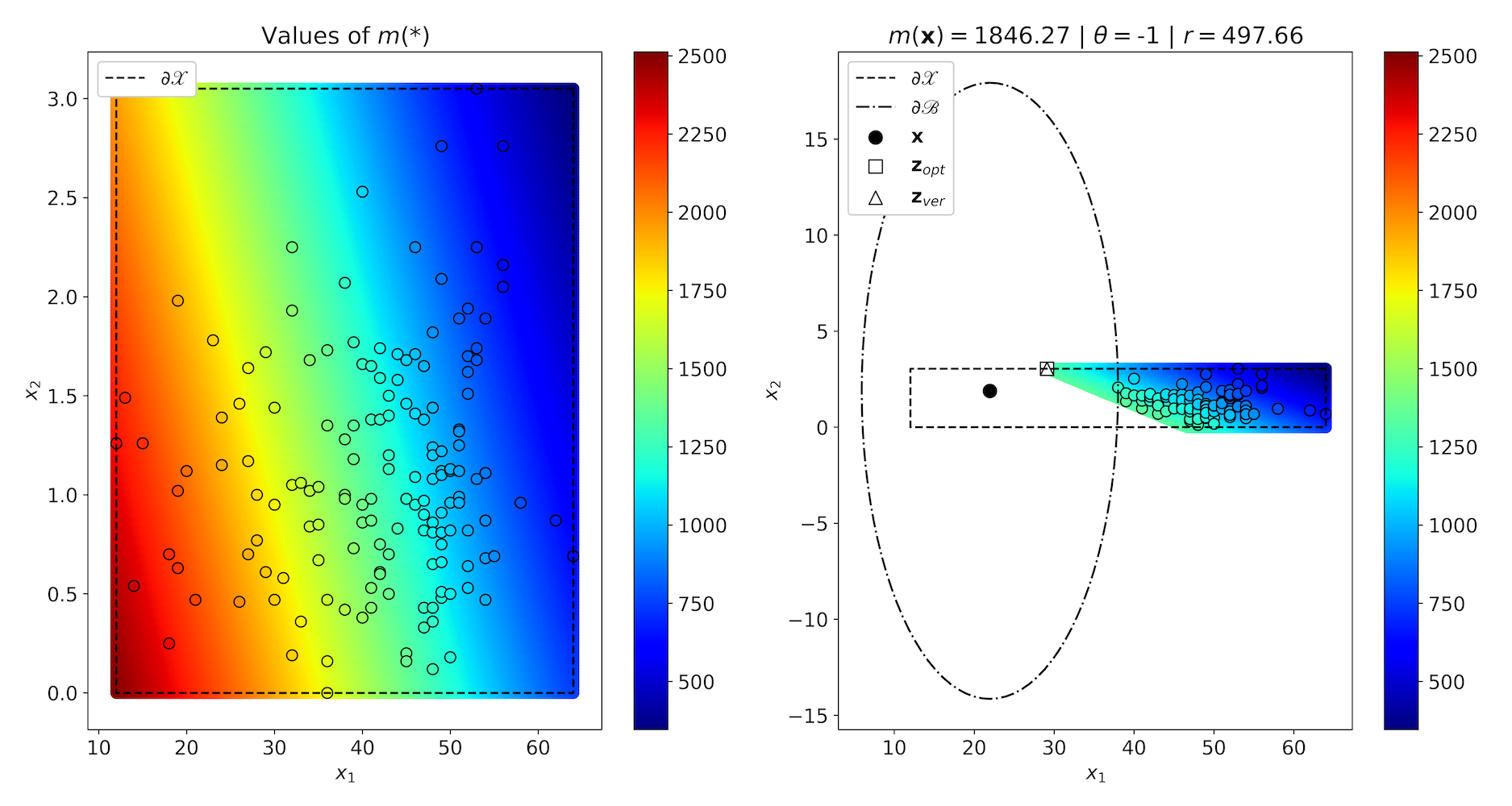}%
\caption{Original and counterfactual points by $\theta=-1$ for the black-box
Cox model trained on the dataset Stanford2}%
\label{f:_stanford2_cox_2}%
\end{center}
\end{figure}

Since examples from the dataset Stanford2 have two features which can be
changed (age $x_{1}$ and T5 mismatch score $x_{2}$), then results of numerical
experiments for this dataset can be visualized, and they shown in Figs.
\ref{stanford2_cox_1}-\ref{f:_stanford2_cox_2}. We again see that points
$\mathbf{z}_{ver}$ and $\mathbf{z}_{opt}$ are close to each other. The same
follows from Table \ref{t:Contraf_Real_Cox} which is similar to Table
\ref{t:Contraf_Syn_Cox}, but contains results obtained for real data. If to
consider values in the last column of Table \ref{t:Contraf_Real_Cox} as the
method accuracy values, then one can conclude that the method provides
outperforming results. This means that the Cox model used as a black-box model
accurately supports the dataset, and the PSO provides a good solution.%

%

Results of experiments with the Cox model trained on datasets Stanford2,
Myeloid-A and Myeloid-B are shown in Table \ref{t:Contraf_Real_Cox}. One can
see that the proposed method also provides exact results for datasets
Myeloid-A and Myeloid-B.%

\begin{table}[tbp] \centering
\caption{Results of numerical experiments for the black-box Cox model trained on real data}%
\begin{tabular}
[c]{cccccccc}\hline
Dataset & $\theta$ & $r$ & $r_{ver}$ & $r_{opt}$ & $\left\Vert \mathbf{z}%
_{ver}-\mathbf{x}\right\Vert _{2}$ & $\left\Vert \mathbf{z}_{opt}%
-\mathbf{x}\right\Vert _{2}$ & $\left\Vert \mathbf{z}_{ver}-\mathbf{z}%
_{opt}\right\Vert _{2}$\\\hline
& $1$ & $198.54$ & $198.54$ & $198.54$ & $0.983$ & $0.983$ & $5.62\times
10^{-7}$\\\cline{2-8}%
Stanford2 & $-1$ & $497.66$ & $497.66$ & $497.66$ & $7.217$ & $7.217$ &
$7.96\times10^{-9}$\\\cline{2-8}
& $1$ & $805.36$ & $805.36$ & $805.36$ & $9.145$ & $9.145$ & $1.03\times
10^{-8}$\\\cline{2-8}
& $-1$ & $186.49$ & $186.49$ & $186.49$ & $1.663$ & $1.663$ & $1.16\times
10^{-8}$\\\hline
& $1$ & $600.05$ & $600.05$ & $600.05$ & $205.937$ & $205.937$ &
$2.36\times10^{-4}$\\\cline{2-8}%
Myeloid-A & $-1$ & $144.24$ & $144.24$ & $144.24$ & $40.383$ & $40.383$ &
$2.15\times10^{-5}$\\\cline{2-8}
& $1$ & $362.00$ & $362.00$ & $362.00$ & $103.535$ & $103.535$ &
$2.16\times10^{-4}$\\\cline{2-8}
& $-1$ & $318.66$ & $318.66$ & $318.66$ & $124.692$ & $124.692$ &
$1.11\times10^{-3}$\\\hline
& $1$ & $57.08$ & $57.08$ & $57.08$ & $28.437$ & $28.437$ & $2.30\times
10^{-5}$\\\cline{2-8}%
Myeloid-B & $-1$ & $421.76$ & $421.76$ & $421.76$ & $126.749$ & $126.749$ &
$5.10\times10^{-4}$\\\cline{2-8}
& $1$ & $206.76$ & $206.76$ & $206.76$ & $260.912$ & $260.912$ &
$2.77\times10^{-3}$\\\cline{2-8}
& $-1$ & $498.91$ & $498.91$ & $498.91$ & $124.941$ & $124.941$ &
$4.85\times10^{-4}$\\\hline
\end{tabular}
\label{t:Contraf_Real_Cox}%
\end{table}%

\subsubsection{The black-box RSF}

\begin{figure}
[ptb]
\begin{center}
\includegraphics[
height=2.527in,
width=4.7089in
]%
{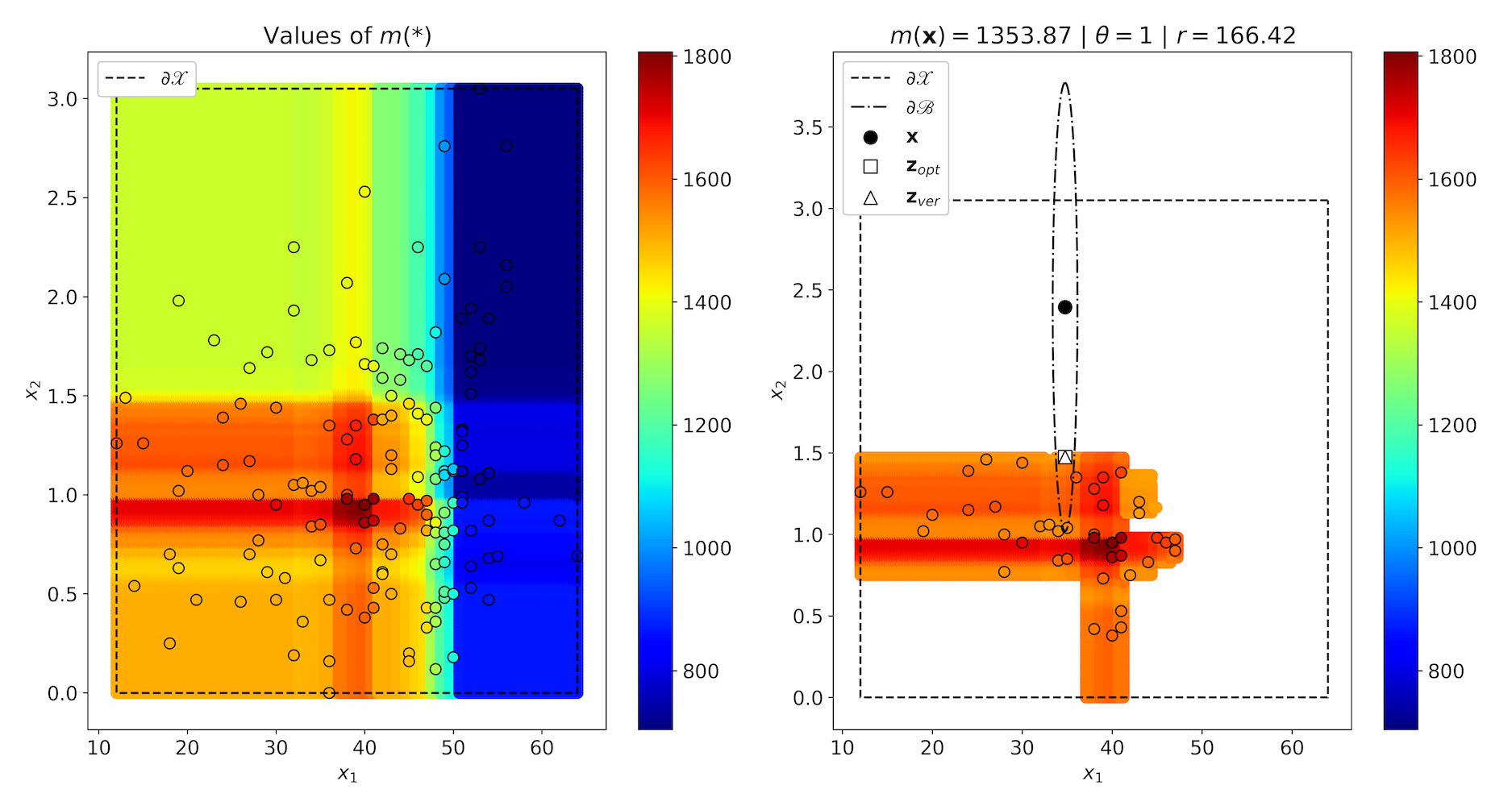}%
\caption{Original and counterfactual points by $\theta=1$ for the black-box
RSF trained on the dataset Stanford2}%
\label{f:real_rsf_1}%
\end{center}
\end{figure}

\begin{figure}
[ptb]
\begin{center}
\includegraphics[
height=2.5235in,
width=4.7003in
]%
{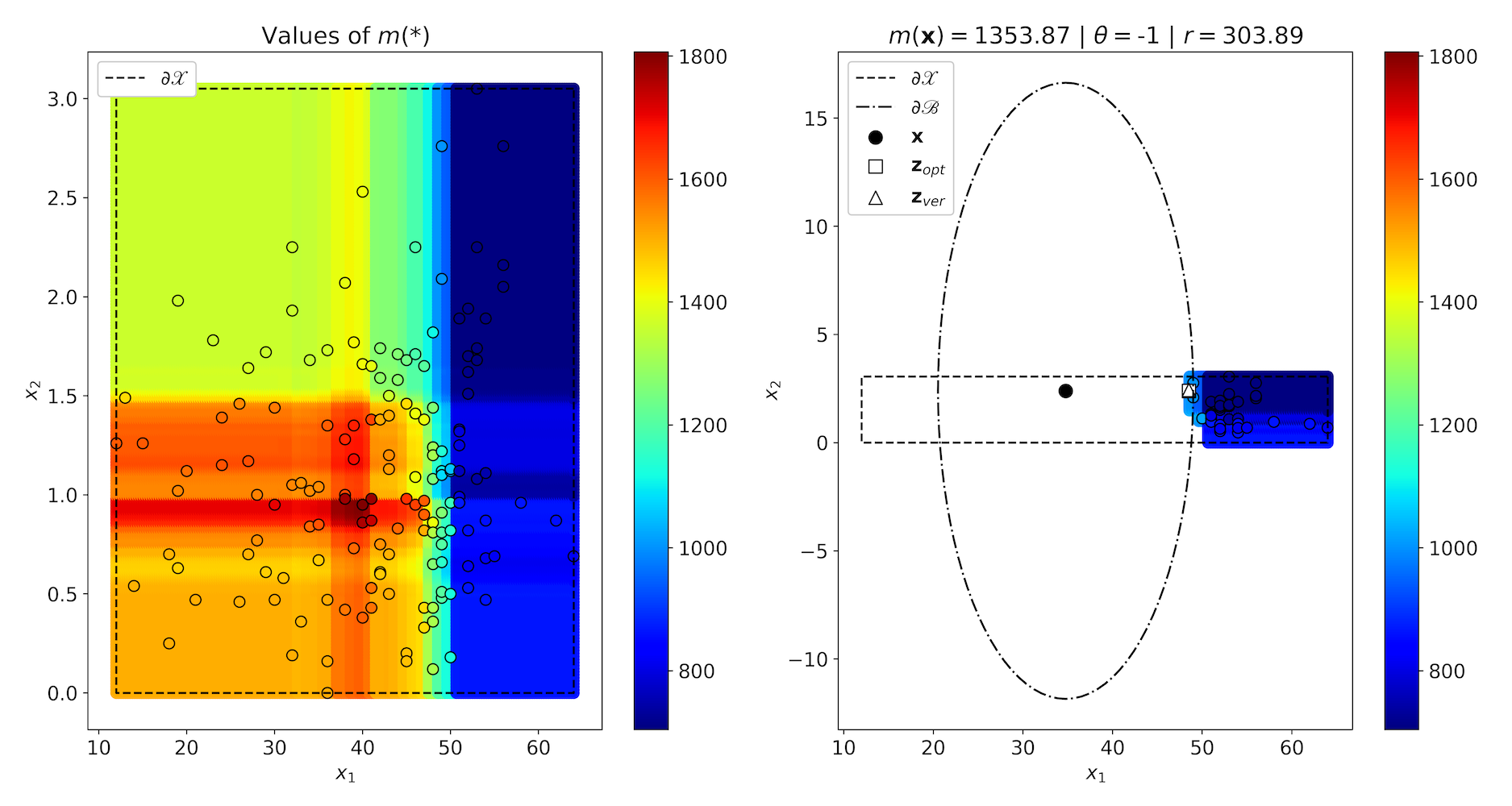}%
\caption{Original and counterfactual points by $\theta=-1$ for the black-box
RSF trained on the dataset Stanford2}%
\label{f:real_rsf_2}%
\end{center}
\end{figure}

Results for the black-box RSF trained on the dataset Stanford2 are given in
Figs. \ref{f:real_rsf_1}-\ref{f:real_rsf_2}. We again see that $\mathbf{z}%
_{ver}$ is close to $\mathbf{z}_{opt}$. Results of experiments with datasets
Stanford2, Myeloid-A and Myeloid-B are shown in Table \ref{t:Contraf_Real_RSF}%
. We again see that values of $A\mathbf{\ }$positive for all datasets. This
means that the PSO gives better results than the method based on generating
the large number of random points.%

%

%

\begin{table}[tbp] \centering
\caption{Results of numerical experiments for the black-box RSF trained on real data}%
\begin{tabular}
[c]{cccccccc}\hline
Dataset & $\theta$ & $r$ & $r_{ver}$ & $r_{opt}$ & $\left\Vert \mathbf{z}%
_{ver}-\mathbf{x}\right\Vert _{2}$ & $\left\Vert \mathbf{z}_{opt}%
-\mathbf{x}\right\Vert _{2}$ & $A$\\\hline
& $1$ & $225.27$ & $236.72$ & $236.72$ & $0.533$ & $0.532$ & $0.0006$%
\\\cline{2-8}%
Stanford2 & $-1$ & $417.90$ & $418.24$ & $418.24$ & $27.544$ & $27.541$ &
$0.0027$\\\cline{2-8}
& $1$ & $166.42$ & $171.66$ & $171.66$ & $0.921$ & $0.920$ & $0.0002$%
\\\cline{2-8}
& $-1$ & $303.89$ & $348.91$ & $348.91$ & $13.738$ & $13.738$ & $0.0002$%
\\\hline
& $1$ & $9.26$ & $10.10$ & $10.10$ & $99.917$ & $99.689$ & $0.2282$%
\\\cline{2-8}%
Myeloid-A & $-1$ & $36.72$ & $37.36$ & $37.36$ & $275.622$ & $274.969$ &
$0.6532$\\\cline{2-8}
& $1$ & $6.73$ & $10.10$ & $10.10$ & $130.504$ & $130.315$ & $0.1888$%
\\\cline{2-8}
& $-1$ & $24.66$ & $25.03$ & $25.03$ & $88.392$ & $86.078$ & $2.3143$\\\hline
& $1$ & $28.52$ & $30.02$ & $30.02$ & $100.882$ & $99.770$ & $1.1120$%
\\\cline{2-8}%
Myeloid-B & $-1$ & $197.97$ & $200.77$ & $198.88$ & $523.077$ & $521.193$ &
$1.8838$\\\cline{2-8}
& $1$ & $2.10$ & $5.54$ & $5.54$ & $123.137$ & $122.755$ & $0.3818$%
\\\cline{2-8}
& $-1$ & $185.27$ & $196.00$ & $192.56$ & $245.575$ & $244.157$ &
$1.4174$\\\hline
\end{tabular}
\label{t:Contraf_Real_RSF}%
\end{table}%

\section{Conclusion}

One of the important difficulties of using the proposed method is to take into
account categorical feature. The difficulty is that the optimization problem
cannot handle categorical data and becomes a mixed integer convex optimization
problem whose solving is a difficult task in general. Sharma et al.
\cite{Sharma-etal-2019} proposed a genetic algorithm called CERTIFAI to
partially cope with the problem and for computing counterfactuals. The same
problem was studied by Russel \cite{Russel-19}. Nevertheless, an efficient
solver for this problem is a direction for further research. The same problem
takes place in using the PSO. There are some modifications of the original PSO
taking into account the categorical and integer features
\cite{Chowdhury-etal-13,Laskari-etal-02,Strasser-etal-16}. However, their
application to the considered explanation problem is another direction for
further research.

We have studied only one criterion for comparison of the SFs of the original
example and the counterfactual. This criterion is the difference of mean
values. In fact, this criterion implicitly defines different classes of
examples. However, other criteria can be applied to the problem and to
separating the classes, for example, difference between values of SFs at time
moment. The study of other criteria is also an important direction for further research.

Another interesting problem is when the feature vector is a picture, for
example, a computer tomography image of an organ. In this case, we have a
high-dimensional explanation problem whose efficient solution is also a
direction for further research.

\section*{Acknowledgement}

This work is supported by the Russian Science Foundation under grant 18-11-00078.


\end{document}